\def\year{2020}\relax
\documentclass[letterpaper]{article} 
\usepackage{aaai20}  
\usepackage{times}  
\usepackage{helvet} 
\usepackage{courier}  
\usepackage[hyphens]{url}  
\usepackage{graphicx} 

\usepackage{subcaption}
\usepackage{amsmath}
\usepackage{enumerate}
\usepackage{multirow}
\usepackage{multicol}

\usepackage{amsfonts}       
\usepackage{nicefrac}       
\usepackage{microtype}      

\DeclareMathOperator*{\argmin}{arg\,min}

\usepackage{graphicx}  
\title{3D Shape Completion with Multi-view Consistent Inference}

\author{\textbf{Tao Hu, Zhizhong Han, Matthias Zwicker}\\ 
	Department of Computer Science\\ 
	University of Maryland, College Park\\
	taohu@cs.umd.edu, h312h@umd.edu, zwicker@cs.umd.edu 
}

\begin{document}
\maketitle
\begin{abstract}
3D shape completion is important to enable machines to perceive the complete geometry of objects from partial observations. To address this problem, view-based methods have been presented. These methods represent shapes as multiple depth images, which can be back-projected to yield corresponding 3D point clouds, and they perform shape completion by learning to complete each depth image using neural networks. While view-based methods lead to state-of-the-art results, they currently do not enforce geometric consistency among the completed views during the inference stage. To resolve this issue, we propose a multi-view consistent inference technique for 3D shape completion, which we express as an energy minimization problem including a data term and a regularization term. We formulate the regularization term as a consistency loss that encourages geometric consistency among multiple views, while the data term guarantees that the optimized views do not drift away too much from a learned shape descriptor. Experimental results demonstrate that our method completes shapes more accurately than previous techniques.
\end{abstract}

\section{Introduction}

Convolutional neural networks have proven highly successful at analysis and synthesis of visual data such as images and videos. This has spurred interest in applying convolutional network architectures also to 3D shapes, where a key challenge is to find suitable generalizations of discrete convolutions to the 3D domain. Popular techniques include using discrete convolutions on 3D grids \cite{3D_ShapeNets}, graph convolutions on meshes \cite{litany2017deformable}, convolution-like operators on 3D point clouds \cite{Atzmon:2018:PCN,Li2018PCNN}, or 2D convolutions on 2D shape parameterizations  \cite{Cohen2018SCNN}. A simple approach in the last category is to represent shapes using multiple 2D projections, or multiple depth images, and apply 2D convolutions on these views. This has led to successful techniques for shape classification \cite{su15mvcnn}, single-view 3D reconstruction \cite{Richter2018MNW}, shape completion \cite{mvcn}, and shape synthesis \cite{Soltani2017S3D}.
One issue in these approaches, however, is to encourage consistency among the separate views and avoid that each view represents a slightly different object. This is not an issue in supervised training, where the loss encourages all views to match the ground truth shape. But at inference time or in unsupervised training, ground truth is not available and a different mechanism is required to encourage consistency.

In this paper, we address the problem of shape completion using a multi-view depth image representation, and we propose a multi-view consistency loss that is minimized during inference. 
We formulate inference as an energy minimization problem, where the energy is the sum of a data term given by a conditional generative net, and a regularization term given by a geometric consistency loss. Our results show the benefits of optimizing  geometric consistency in a multi-view shape representation during inference, and we demonstrate that our approach leads to state-of-the-art results in shape completion benchmarks. In summary, our contributions are as follows: 

\begin{enumerate}[i)]
	\item We propose a multi-view consistency loss for 3D shape completion that does not rely on ground truth data. 
	\item We formulate multi-view consistent inference as an energy minimization problem including our consistency loss as a regularizer, and a neural network-based data term.
	\item We show state-of-the-art results in standard shape completion benchmarks, demonstrating the benefits of the multi-view consistency loss in practice. 
\end{enumerate}

\begin{figure*}[t]
	\centering
	\includegraphics[width=\linewidth]{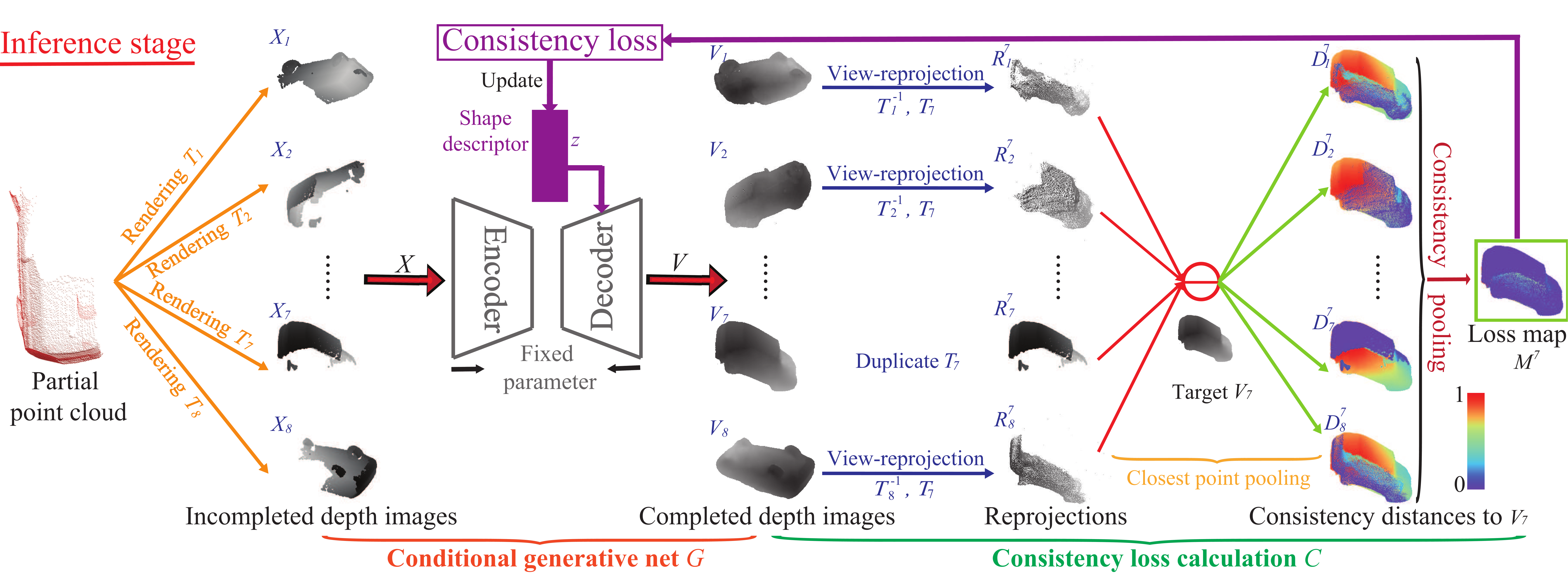}
	\caption{Overview of the multi-view consistent inference for 3D shape completion. Given a partial point cloud as input, we first render multiple incomplete views $X$, which form our shape representation of the incomplete input. To perform inference, we apply a conditional generative network $G$ to generate completed depth images $V$ based on a shape descriptor $z$ conditioned on $X$. As a key idea, we design our consistency loss $C$ to evaluate the geometric consistency among $V$. Intuitively, for all pixels in all views $V_t$ we find the distance to their approximate closest neighbor in the other views $V_s$, and sum up these distances to form $C$. Specifically, for each target view (e.g. $V_7$ in the figure) we reproject all completed depth images $V_s$ according to the pose of $V_7$, which leads to reprojection maps denoted $R^7_s$. Then we compute consistency distances, denoted $D^7_s$, for each reprojection map $R^7_s$ and the target $V_7$ via a pixel-wise closest point pooling operation. Finally, a consistency pooling operator aggregates all consistency distances $D^7_s$ into a loss map $M^7$. In inference, we minimize all loss maps as a function of the shape descriptor $z$.
	}
	\label{fig:overview}
\end{figure*}

\section{Related Work}
\label{related_work}
\noindent\textbf{3D shape completion.} Different 3D shape representations have been applied in 3D shape completion, such as voxels, point clouds, and multiple views. Voxel-based representations are widely used in shape completion with 3D CNN, such as 3D-Encoder-Predictor CNNs \cite{epn3d} and encoder-decoder CNN for patch-level geometry refinement \cite{high_reso}. However, computational complexity grows cubically as the voxel resolution increases, which severely limits the completion accuracy. To address this problem, several point cloud-based shape completion methods \cite{fc_Achlioptas2018LearningRA,folding,ref_pcn} have been proposed. The point completion network (PCN)~\cite{ref_pcn} is a current state-of-the-art approach that extends the PointNet architecture~\cite{pointnet} to provide an encoder, followed by a multi-stage decoder that uses both fully connected~\cite{fc_Achlioptas2018LearningRA} and folding layers~\cite{folding}.
The output point cloud size in these methods is fixed, however, to small numbers like 2048 \cite{folding}, which often leads to the loss of detail. View-based methods resolve this issue by completing each rendered depth image \cite{mvcn} of the incomplete shape, and then back-projecting the completed images into a dense point cloud. By leveraging state-of-the-art image-to-image translation networks \cite{pix2pix2016}, MVCN \cite{mvcn} completed each single view with a shape descriptor which encodes the characteristics of the whole 3D object to achieve higher accuracy. 
However, view-based methods fail to maintain geometric consistency among completed views during inference. Our approach resolves this issue using our novel multi-view consistent inference technique.

\noindent\textbf{Multi-view consistency.} One problem of view-based representation is inconsistency among multiple views. Some researchers presented a multi-view loss to train their network to achieve consistency in multi-view representations, like discovering 3D keypoints \cite{keypoint} and reconstructing 3D objects from images ~\cite{lin2018learning,LiPZR18,mvcTulsiani18,JiangSQJ18,Khot2019LearningUM}. With differentiable rendering \cite{lin2018learning,mvcTulsiani18}, the consistency distances among different views can be leveraged as 2D supervision to learn 3D shapes in their networks. However, these methods can only guarantee consistency for training data in training stage. Different from these methods, with the help of our novel energy optimization and consistency loss implementation, our proposed method can improve geometric consistency on test data directly during the inference stage.


\begin{figure*}[t]
	\centering
	\begin{minipage}{.4\textwidth}
		\centering
		\includegraphics[width=\linewidth]{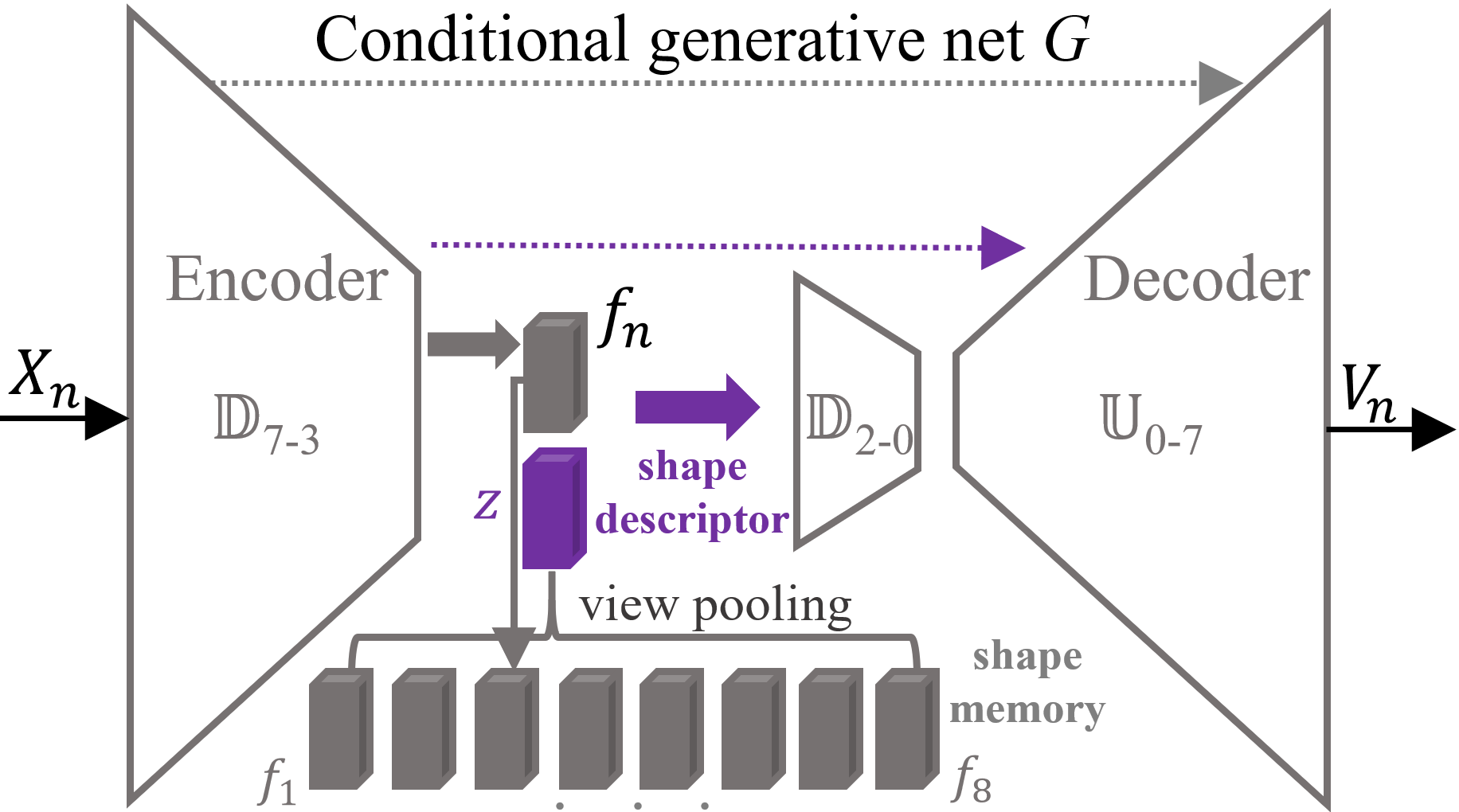}	
		\captionof{figure}{Net structure.}
		\label{fig:net}
	\end{minipage}%
	\begin{minipage}{.58\textwidth}
		\centering
		\includegraphics[width=\linewidth]{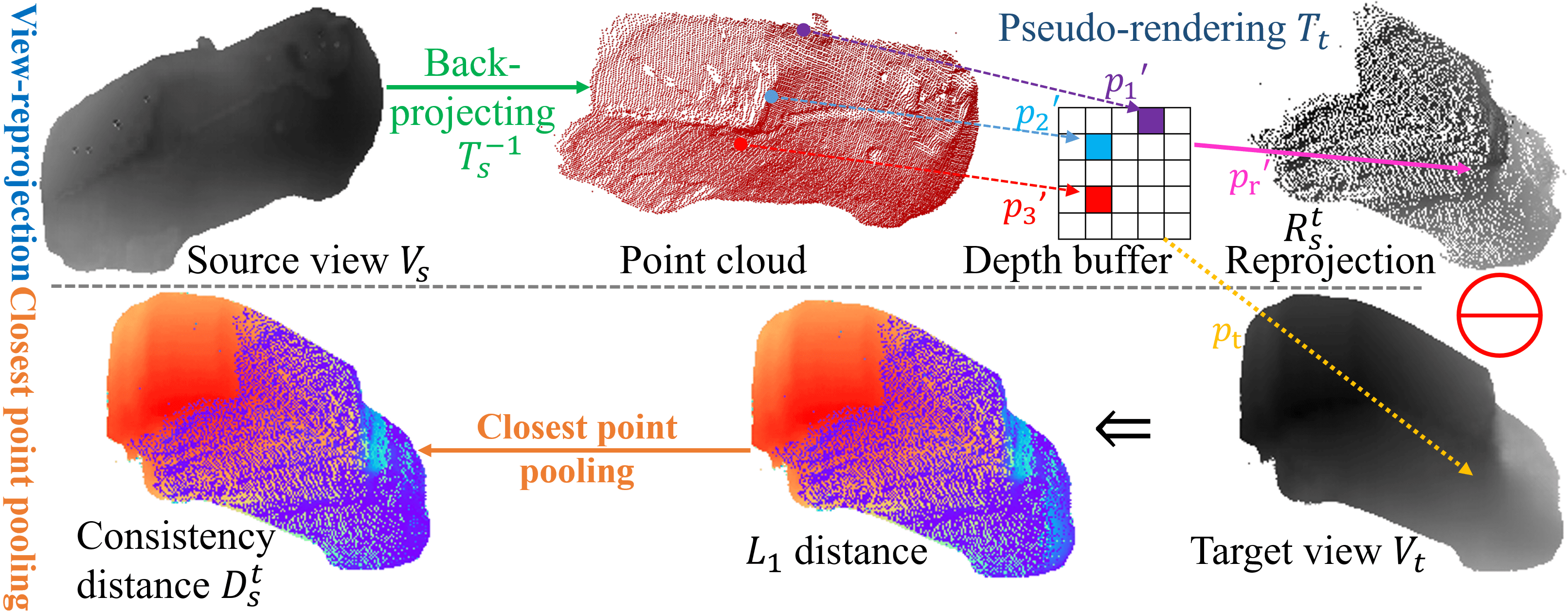}
		\captionof{figure}{Methods to calculate consistency distance.}
		\label{fig:pipeline}
	\end{minipage}
\end{figure*}

\section{Multi-view Consistent Inference}
\label{methods}

\noindent\textbf{Overview. }The goal of our method is to guarantee multi-view consistency in inference, as shown in the overview in Fig.~\ref{fig:overview}. Our method starts from converting partial point clouds to multi-view depth image representations by rendering the points into a set of incomplete depth images $X=\{X_1,\dots,X_8\}$ from a number of fixed viewpoints. In our current implementation, we use eight viewpoints placed on the corners of a cube. Our approach builds on a conditional generative net $G({z}; X)$ which is trained to output completed depth images $V$ by estimating a shape descriptor ${z}$ conditioned on a set of incomplete inputs $X$. We obtain the conditional generative net in a separate, supervised training stage. During inference, we keep the network weights fixed and optimize the shape descriptor ${z}$ to minimize an energy consisting of a consistency loss, which acts as a regularizer, and a data term. On the one hand, the consistency loss $C(V) = C(G({z}; X))$ quantifies the geometric consistency among the completed depth images $V$. On the other hand, the data term encourages the solution to stay close to an initially estimated shape descriptor $\mathring{z}$. 
This leads to the following optimization for the desired shape descriptor $z^*$:

\begin{equation}
\begin{split}
z^* & =  \argmin_{{z}} C(G({z};X)) +\mu||G({z};X)-G(\mathring{z};X)|| \\
& =\mathcal{L}_{con}(z) +  \mu\mathcal{L}_{gen}(z),
\label{eq:overview}
\end{split}
\end{equation}

where $\mu$ is a weighting factor, and we denote $Y=G(\mathring{z};X)$ and $V=G({z};X)$ as the initially estimated completed depth images and optimized completed depth images in inference, respectively. In addition, we will formulate the regularization term and data term as multi-view consistency loss $\mathcal{L}_{con}(z)$ and generator loss $\mathcal{L}_{gen}(z)$ in Section `Consistency Loss'.



\noindent\textbf{Conditional generative net. }The conditional generative net $G({z};X)$ is built on the structure of multi-view completion net \cite{mvcn}, as shown in Fig.~\ref{fig:net}, which is an image-to-image translation architecture applied to perform depth image completion for multiple views of the same shape. 
We train the conditional generative net following a standard conditional GAN approach \cite{GAN}. To share information between multiple depth images of the same shape,
our architecture learns a shape descriptor ${z}$ for each 3D object by pooling a so-called shape memory consisting of $N$ feature maps $f_n, n \in [1,N=8]$ from all views of the shape. The network $G$ consists of 8 U-Net modules, and each U-Net module has two submodules, Down and Up, so there are 8 Down submodules ($\mathbb{D}_{7-0}$) in the encoder and 8 Up submodules ($\mathbb{U}_{0-7}$) in the decoder. Down submodules consist of the form Convolution-BatchNorm-ReLU\cite{Ioffe2015BatchNA,Nair2010RectifiedLU}, and Up submodules ($\mathbb{D}_{0-7}$) consist of the form UpReLU-UpConv-UpNorm. The shape memory is the feature map after the third Down submodule ($\mathbb{D}_{3}$) of the encoder. More details can be found in \cite{pix2pix2016,mvcn}. 

In inference, we optimize the shape descriptor $z$ of $G({z};X)$ given test input $X$. We first get an initial estimation of the shape descriptor $\mathring{z}$ for each test shape by running the trained model once, and initialize $z$ with  $\mathring{z}$. During inference the other parameters of $G$ are fixed. 


\section{Consistency Loss}
\label{sec:multi_viewloss}	
Our consistency loss is based on the sum of the distances between each pixel in the multi-view depth map and its approximate nearest neighbor in any of the other views. In this section we introduce the details of the multi-view consistency loss calculation following the overview in Fig.~ \ref{fig:overview}. For all views $V_t$, we first calculate pairwise per-pixel consistency distances $D_s^t$ to each other view $V_s$, that is, per-pixel distances to approximate nearest neighbors in view $V_s$.
We then perform consistency pooling, which for each view $V_t$ provides the consistency distances over all other views (as opposed to the initial pairwise consistency distances between two of views). We call these the loss maps $M^t$. The final consistency loss is the sum over all loss maps.

\subsection{Pairwise Consistency Distances}

Given a source view $V_s$ and a target view $V_t$, we calculate the consistency distance $D_s^t$ between $V_s$ and $V_t$ by view-reprojection  and closest point pooling, where $V_t, V_s \in \mathbb{R}^{H \times W}$ and $H\times W$ is the image resolution. Specifically, view-reprojection transforms the depth information of source $V_s$ to a reprojection map $R_s^t$ according to the transformation matrix of the target $V_t$. Then, closest point pooling further produces the consistency distance $D_s^t$ between $R_s^t$ and $V_t$. Fig.~ \ref{fig:pipeline} shows the pipeline, where the target view is $V_7$ and the source view is $V_2$. 
In the following, we denote a pixel on source view as $p_i=[{u}_i, {v}_i,{d}_i]$, where $u_i$ and $v_i$ are considered pixel coordinates, its back-projected 3D point as $P_i=[\hat{x}_i,\hat{y}_i,\hat{z}_i]$, and the reprojected pixel on reprojection map $R_s^t$ as ${p'_i}=[{u}'_i,{v}'_i, {d}'_i]$, where ${d}_i=V_s[{u}_i,{v}_i]$ and ${d}'_i=R_s^t[{u}'_i,{v}'_i]$ are the depth values at the location $[{u}_i,{v}_i]$ and $[{u}'_i,{v}'_i]$, respectively.


\noindent \textbf{View-reprojection}. The view-reprojection operator back-projects each point $p_i=[{u}_i,{v}_i,{d}_i]$ on $V_s$ into the canonical 3D coordinates as $P_i=[\hat{x}_i,\hat{y}_i,\hat{z}_i]$ via 

\begin{equation}
{P}_i=\Re^{-1}_s(K^{-1}{p}_i-\tau_s)  \quad \forall i, 
\label{eq:backproject}
\end{equation}
where $K$ is the intrinsic camera matrix, and $\Re_s$ and $\tau_s$ are the rotation matrix and translation vector of view $V_s$ respectively. This defines the relationship between the view $V_s=\{p_i\}$ and its back-projected point cloud $\{P_i\}$. We use $T_s$ to denote the transformation matrix of $V_s$, which contains the pose information, such that $T_s = (\Re_s,\tau_s)$.
Then, we transform each 3D point $P_i$ in the point cloud into a pixel $p'_i=[{u}'_i, {v}'_i, {d}'_i]$ on the reprojection map $R^t_s$ as

\begin{equation}
{p}'_i=K(\Re_t {P}_i + \tau_t)  \quad  \forall i.
\label{eq:project}
\end{equation}

Eq.~(\ref{eq:backproject}) and Eq.~(\ref{eq:project}) illustrate that we can transform the depth information of source view $V_s$ to reprojection map $R^t_s$, which has the same pose with the target view $V_t$. However, due to the discrete grid of the depth images, different points $P_i$ in the point cloud may be projected to the same pixel $[u',v']$ on the reprojection map $R_s^t$ when using Eq.~(\ref{eq:project}), like ${p'_1}=[{u}',{v}', {d}'_1], {p'_2}=[{u}',{v}', {d}'_2], {p'_3}=[{u}',{v}', {d}'_3]$ in Fig.~ \ref{fig:pipeline}. In fact, all the $\{{p'_1},{p'_2},{p'_3}\}$ are projected to the same pixel ${p'_r}$ on $R_s^t$, and the corresponding point on the target view $V_t$ is $p_t$.
To alleviate this collision effect, we implement a pseudo-rendering technique similar to ~\cite{lin2018learning}. Specifically, for each pixel on $R_s^t$, a sub-pixel grid with a size of ($U \times U$) is presented to store multiple depth values corresponding to the same pixel, so the reprojection is $ R^t_s \in \mathbb{R}^ {H \times U \times W \times U}$. 

\noindent \textbf{Closest point pooling}. The closest point pooling operator computes the consistency distance between reprojection $R^t_s$ and target view $V_t$. First, we also upsample $V_t$ to $ \mathbb{R}^ {H \times U \times W \times U}$ by repeating each depth value into a $U\times U$ sub-pixel grid. Then, we calculate the element-wise $L_1$ distance between $R^t_s$ and the upsampled $V_t$. Finally, we perform closest point pooling to extract the minimal $L_1$ distance in each sub-pixel grid using min-pooling with a $U\times U$ filter and a stride of $U\times U$. This provides the consistency distance $D^t_s$ between source view $V_s$ and target view $V_t$, where $D^t_s \in \mathbb{R}^ {H \times W}$. The consistency distance $D^t_s$ is shown in Fig.~\ref{fig:pipeline}, where $t=7, s=2$. Note that we directly take the $t$th input view $X_t$ as the reprojection $R^t_t$ when $t=s$, since the incomplete input $X_t$ also provides some supervision.

Note that some consistency distances in $D^t_s$ may be large due to noisy view completion or self-occlusion between the source and target views, and these outliers interfere with our energy minimization. 
Therefore, we perform outlier suppression by ignoring consistency distances above a threshold of $2.5\%$ of the depth range (from the minimum to the maximum depth value of a model).

\subsection{Consistency Distance Aggregation by Consistency Pooling}
\begin{figure*}[t]
	\centering
	\begin{minipage}{.57\textwidth}
		\centering
		\includegraphics[width=\linewidth]{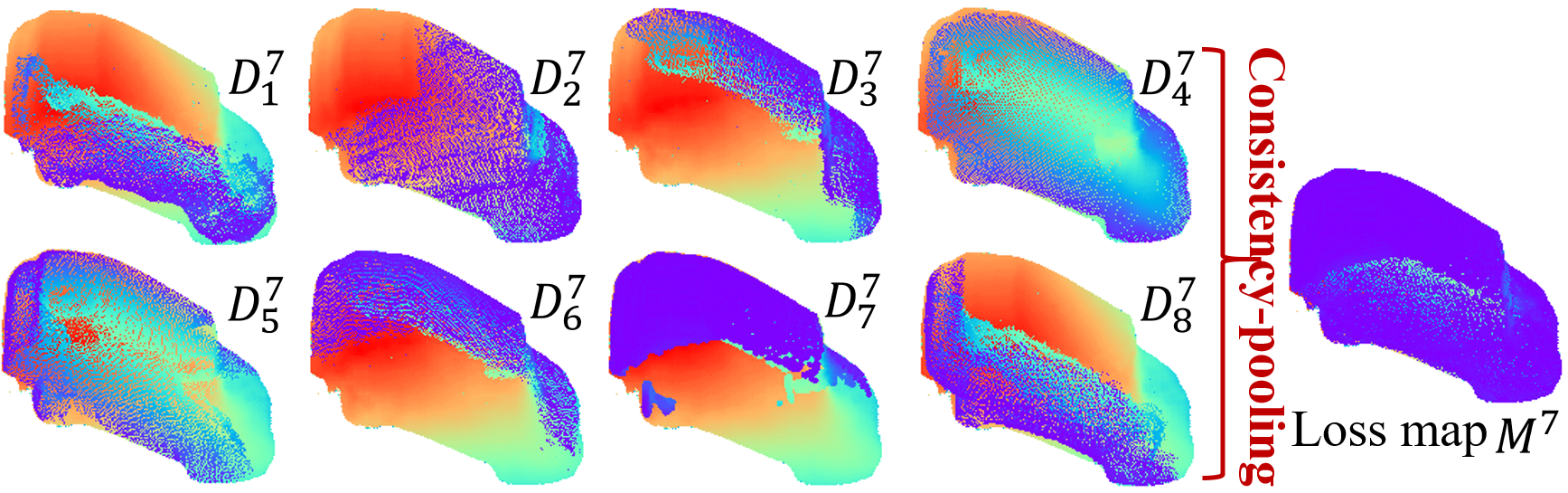}
		\captionof{figure}{Consistency pooling with respect to $V_7$.}
		\label{fig:consist_pooling}
	\end{minipage}%
	\begin{minipage}{.42\textwidth}
		\centering
		\includegraphics[width=\linewidth]{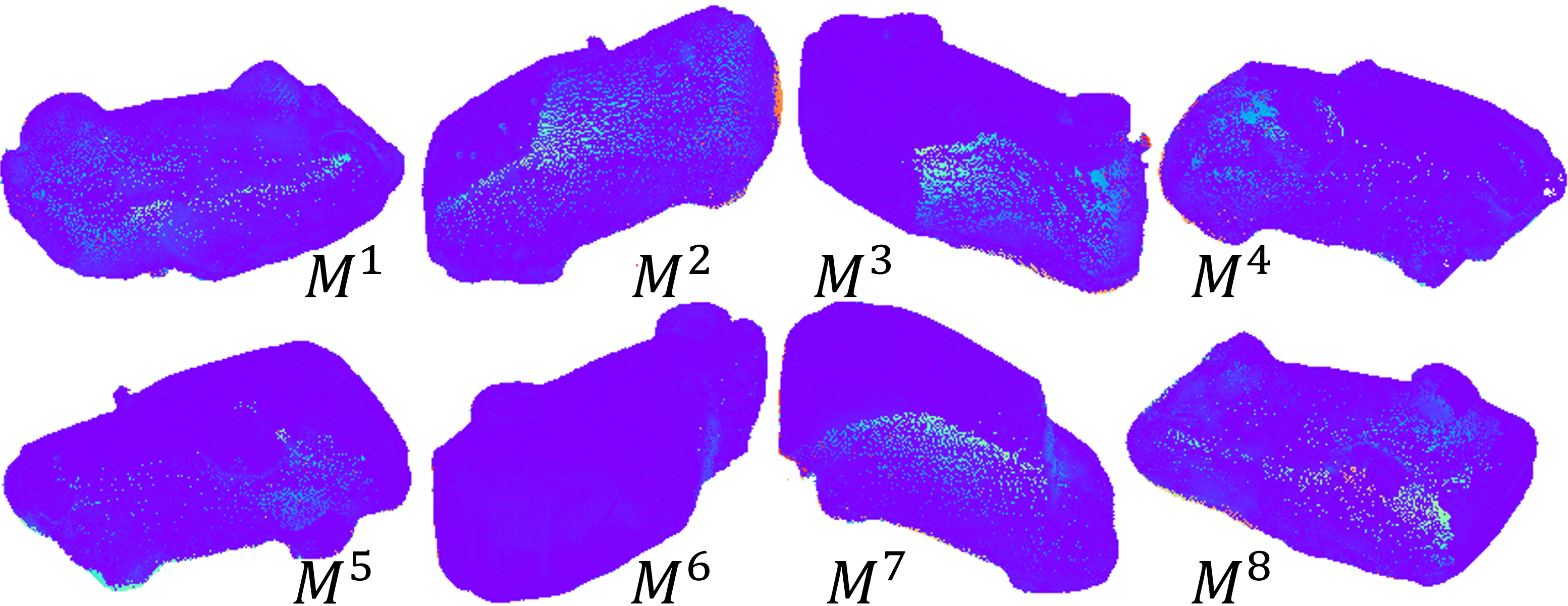}
		\captionof{figure}{Eight loss maps of a 3D model.}
		\label{fig:8nearestmaps}
	\end{minipage}
\end{figure*}

Given a target view $V_t$, we get all the consistency distances $D_t^s$ between $V_t$ and all the other $N$ source views $V_s$, as shown in Fig.~ \ref{fig:consist_pooling}, where $t=7, N=8$, and we use the same colorbar with Fig.~\ref{fig:overview}.  Obviously, different source views $V_s$ cover different parts of the target view $V_t$, which leads to different consistency distances in $D_s^t$. For example, the red parts on each $D_s^t$ in Fig.~\ref{fig:consist_pooling} indicate that they can not be well inferred from the source view, so these parts are not helpful for the optimization of the target view. 

By extracting the minimum distance between $V_t$ and the reprojections from all other views, we cover the whole $V_t$ with the closest points to it and we obtain the loss maps $M^t$. In our pipeline, we implement this efficiently using a consistency pooling operator defined as,
\begin{equation}
M^t(x,y)=  \min \limits_{j \in [1,J]} D^t_j(x,y), 
\label{eq:limi2}
\end{equation}	
where $M^t \in \mathbb{R}^{H \times W}$, $x \in [1,H], y \in [1,W]$, and $J$ is the number of views in pooling. We use $J\leq N$ to make it possible to restrict pooling to a subset of the views (see Section `Experiments' for an evaluation of this parameter). This is illustrated using $M^7$ as an example in Fig.~\ref{fig:consist_pooling}. Fig.~\ref{fig:8nearestmaps} shows all the consistency loss maps to each target view.




\subsection{Loss Function}
\label{sec:loss}
Our multi-view consistent inference aims to maximize the depth consistency across all views by optimizing the shape descriptor $z$ of a 3D model. Therefore, the consistency loss $\mathcal{L}_{con}(z)$ to the whole 3D model takes the loss maps for all target views,

\begin{equation}
\mathcal{L}_{con}(z)= C(G({z};X)) =\frac{1}{N \times H \times W}\sum_{t=1}^{N}  \sum_{x=1}^{H}\sum_{y=1}^{W}M^t(x,y),
\label{eq:limi}
\end{equation}


where $N$ is the number of views and $X$ is the input set of incomplete depth images of the 3D model.

In Eq.~(\ref{eq:overview}), we also have a data term to keep $z$ close to the initial estimation $\mathring{z}$ during inference. We call this the generator loss $\mathcal{L}_{gen}$, which aims to prevent the completed depth images drifting away from the prior learned from the training data:
\begin{equation}
\mathcal{L}_{gen}(z)= \|G({z};X)-G(\mathring{z};X)\|,
\label{eq:eq_l2}
\end{equation}

where $X$ is the input, $Y=G(\mathring{z};X)$ and $V=G({z};X)$ are the initially estimated outputs and optimized outputs respectively, and $X, Y, V \in \mathbb{R}^{N\times H\times W}$. In summary, the overall loss function in inference $\mathcal{L}(z)$ is
\begin{equation}
\mathcal{L}(z)=  \mathcal{L}_{con}(z) +  \mu\mathcal{L}_{gen}(z),
\label{eq:eq_loss_all}
\end{equation}

where $\mu$ is a weighting factor. We optimize the shape descriptor $z$ for 100 gradient descent steps, and we take $z$ with the smallest consistency loss in the last 10 steps as $z^*$. It should be mentioned that since the gradients of $z$ are small, we use a large learning rate of 0.2.

\section{Experiments}
\label{experiments}
Our method is built on MVCN~\cite{mvcn}, a state-of-the-art view-based shape completion method. To fairly evaluate the improvements over MVCN directly, we use the same pipeline a MVCN to generate training and test depth images, where each 3D object is represented by $N=8$ depth maps with a resolution of $256 \times 256$. We take 3D models from ShapeNet \cite{shapenet}. 
Initially, we set $J=N$ in Eq.~\ref{eq:limi} to conduct consistency pooling in the following experiments. 
In addition, we use the same training dataset and hyperparameters with \cite{mvcn} to train the network, and the same test dataset with \cite{mvcn,ref_pcn} to evaluate our methods with Chamfer Distance (CD) \cite{ref_cd}. 


\subsection{Analysis of the Objective Function}
We test different objective functions in Eq.~(\ref{eq:eq_loss_all}) to justify the effectiveness of our methods. Table~\ref{tab:loss_func} shows the quantitative effects of these variations. The experiments are conducted on 100 3D airplane models (besides test dataset or training dataset), which are randomly selected under the constraints that the average CD is close to that of the test dataset in \cite{mvcn}. We change the weighting factor $\mu$ between $\mathcal{L}_{con}(z)$ and $\mathcal{L}_{gen}(z)$, and different distance functions in $\mathcal{L}_{gen}(z)$ (using $L_1$ or $L_2$). When $\mu=0$, only $\mathcal{L}_{con}(z)$ is used in loss function. According to the comparison, we select $L_2$ distance to calculate generator loss, and set $\mu=1$ in the following experiments.

\begin{table}[!htb]
	\centering
	\caption{Chamfer distance over different loss functions in Eq.~\ref{eq:eq_loss_all}. CD is  multiplied by $10^3$. }
	\label{tab:loss_func}		
	\begin{tabular}{c|ccccc|c}
		\hline 
		{$\mu$ } & $\mu=0.1$ &$\mu=1$&$\mu=2$&$\mu=5$&$\mu=10$&$\mu=0$\\
		\hline
		$L_1$& 5.228 & 5.160 & \textbf{5.129} & 5.160 & 5.155 & \multirow{2}*{6.383}\\
		$L_2$& 5.362 & \textbf{5.110} & 5.136 & 5.135 & 5.175&\\
		\hline
	\end{tabular}
	\vspace{-0.1in}		
\end{table}


\subsection{The Size of Depth-buffer in Pseudo-rendering}
\begin{figure*}[ht]
	\centering
	\includegraphics[width=\linewidth]{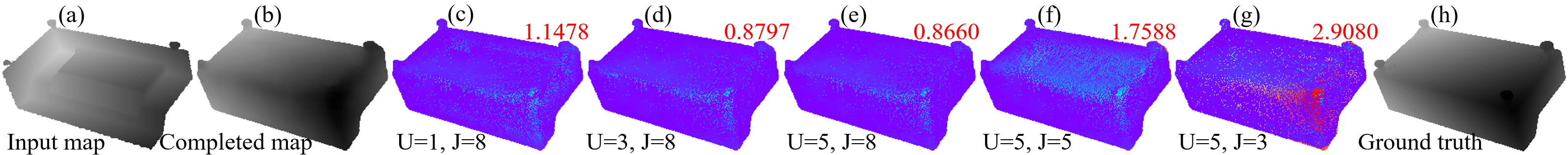}
	\caption{Consistency loss maps over different depth-buffer sizes ($U$) and numbers of views in consistency pooling $J$. (c) to (g) are the consistency loss maps, where the values of the consistency loss (scaled with 100) are marked in red. We use the same colorbar with Fig.~\ref{fig:overview}.}
	\label{fig:uv}
\end{figure*}

As mentioned above, we use a depth-buffer in pseudo-rendering, and the depth-buffer size is $U \times U$. Obviously, a bigger buffer means less collisions in pseudo-rendering, which further makes the reprojection more accurate. The average CD is lower when we increase the size of the depth-buffer, as shown in Table~\ref{tab:uv} (a), where the experiments are conducted on two categories of the test dataset. From the loss maps in Fig.~\ref{fig:uv} (c) to (e), given $J=8$ in consistency pooling Eq.~(\ref{eq:limi}), the consistency loss goes smaller when we increase $U$. This is because the closest points (reprojected from the other 8 views) to the target view are more accurate. We also see less noisy points (brighter ones) in Fig.~\ref{fig:uv} (e). 

\begin{table}[!htb]
	\raggedleft	
	\caption{The effects of depth-buffer sizes $U$ (a) and numbers of views $J$ (in Eq.~\ref{eq:limi2}) in consistency pooling (b). CD is multiplied by $100$. }
	\label{tab:uv}
	\begin{subtable}[c]{.48\linewidth}				
		\begin{tabular}{c|c|c}
			{U} & \multicolumn{2}{c}{Average CD}\\
			\hline
			&Table & Sofa \\
			$U=1$ & 	0.8876	 & 0.8440 \\
			$U=3$	 & 0.8830	 & 0.8421 \\
			$U=5$  & 	\textbf{0.8754} & 	\textbf{0.8394} \\			
			\hline	
		\end{tabular}	
		\caption{}
	\end{subtable}%
	\hfill
	\begin{subtable}[r]{.48\linewidth}
		\raggedright		
		\begin{tabular}{c|c|c}
			{J} & \multicolumn{2}{c}{Average CD}\\
			\hline
			&Table & Sofa \\
			$J=3$ & 	0.8810	 & 0.8484 \\
			$J=5$	 & 0.8764	 & 0.8410 \\
			$J=8$  & 	\textbf{0.8754} & 	\textbf{0.8394} \\			
			\hline	
		\end{tabular}	
		\caption{}	
	\end{subtable}
\end{table}

\subsection{The Number of Views in Consistency Pooling}
In this part, we analyze the effects of varying the number of views $J$ in consistency pooling. As shown in Fig.~ \ref{fig:consist_pooling}, more views mean a bigger coverage over the target view and a smaller consistency loss. Given a depth-buffer size of $5\times 5$, Fig.~\ref{fig:uv} (e) to (g) show that the consistency loss increases when $J=3$ or $J=5$, and we also find more noisy points in these loss maps Fig.~\ref{fig:uv} (f) and (g).

\subsection{Comparison with Direct Optimization Method}

Our multi-view consistent inference can also be used to optimize completed depth maps directly without the conditional generative net $G$. We call this direct optimization on depth maps, and in this part, we compare our methods with direct optimization. In fact, direct optimization only contains the \textit{Consistency loss calculation \textit{C}} part in Fig. ~\ref{fig:overview}. Each depth map will be a trainable tensor. We first initialize the tensors with the completed views $V_n, n\in [1,8]$, and then update these tensors by minimizing the consistency loss in Eq.~ \ref{eq:eq_loss_all}. We use $L_2$ distance to calculate $\mathcal{L}_{gen}(z)$, $\mu=1$, and the learning rate is 0.0006, which produces the best results for direct optimization.
\begin{figure*}[h]
	\centering
	\includegraphics[width=\linewidth]{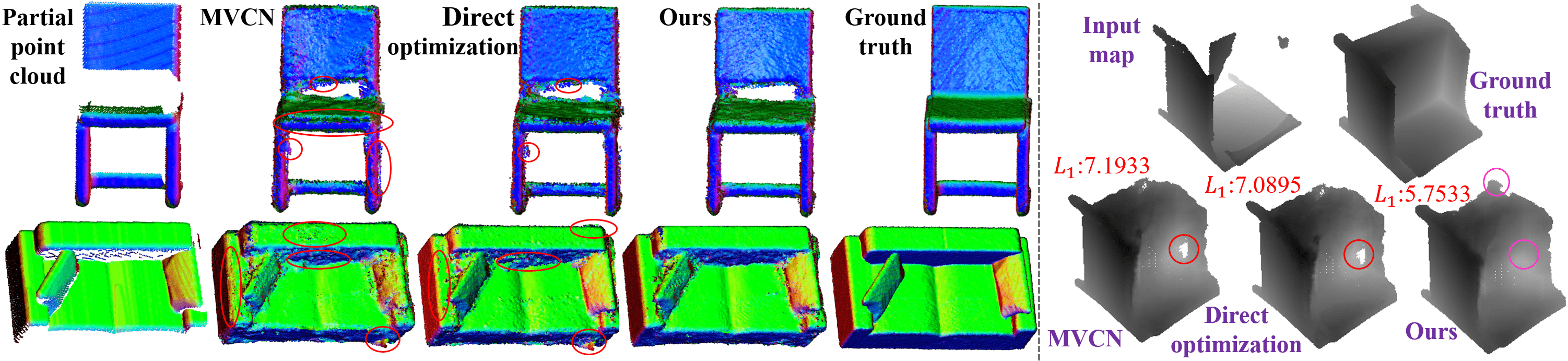}
	\caption{Comparisons between direct optimization and our methods on optimizing point clouds (left) and depth maps (right). The normals of the point clouds are shown.} 
	\label{fig:cmp_direct}
\end{figure*}
Fig.~\ref{fig:cmp_direct} shows the comparisons. Here we color-code the normals of the completed point clouds, which are estimated using a k-d tree search algorithm with a search radius of 0.5 and a maximum number of neighbors of 30. 
Compared with direct optimization, our method performs better. For example, in terms of optimizing point clouds, we can smooth the surface, like the seat of the chair, and remove some outliers. As for completing depth maps, our method can fill a hole appearing in MVCN \cite{mvcn} and even add the missing leg, where the $L_1$ distances to the ground truth are marked in red. 

Though the direct optimization method can also refine the point clouds of MVCN,  it does not perform well in removing outliers on point clouds (left) or completing a depth map (right) in Fig.~\ref{fig:cmp_direct}. The reason is that direct optimization does not have any knowledge to distinguish shape and background from a depth map, which means that for pixels in a hole, direct optimization does not know whether they belong to a hole of the shape or the background. 
However, with the knowledge of shape completion learned in the conditional generative net $G$, our method completes shapes better.

\subsection{Intermediate Results and Convergence}
\begin{figure}[h]
	\centering
	\includegraphics[width=\linewidth]{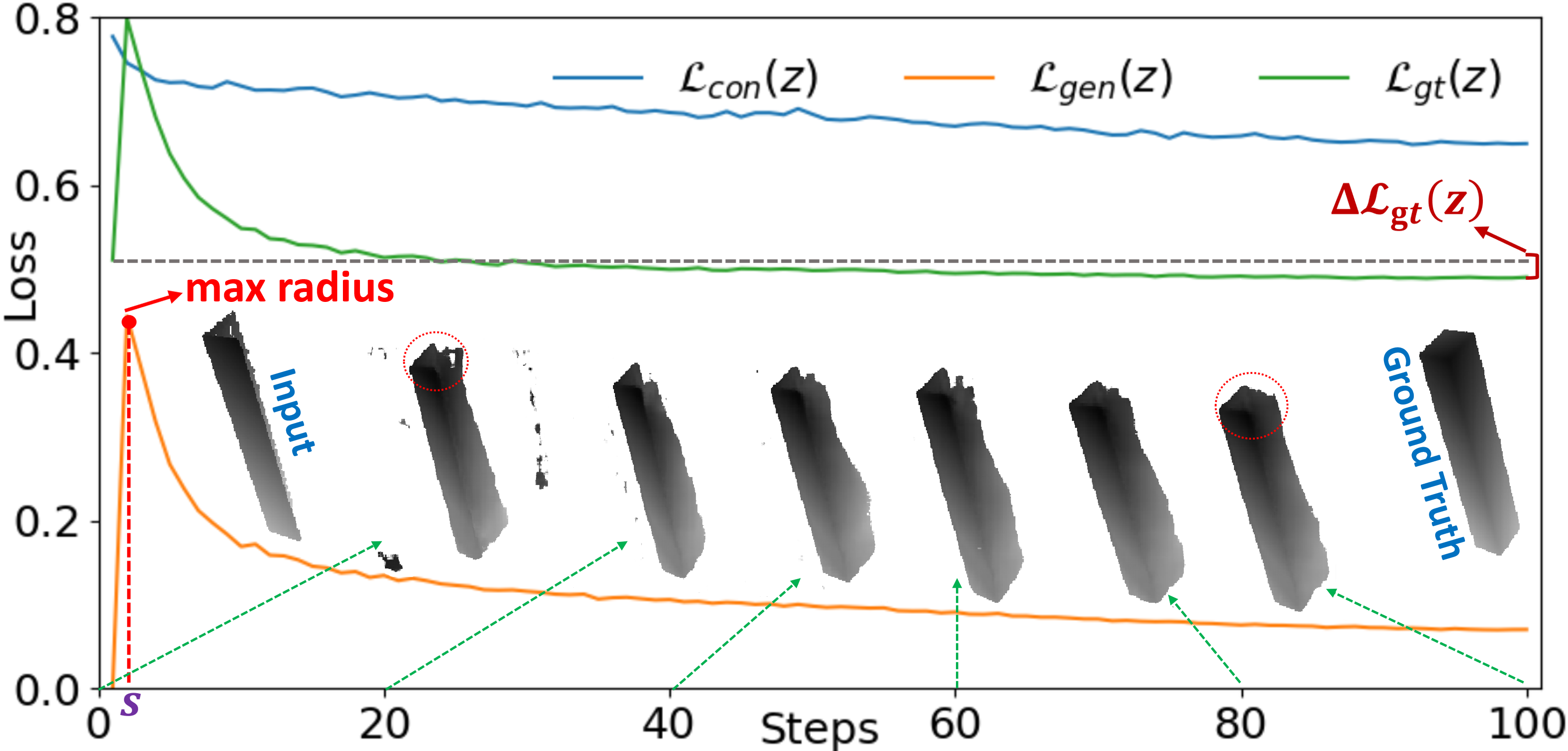}
	\caption{Consistent inference optimization (loss vs steps).}
	\vspace{-0.1in}
\label{fig:curve}
\end{figure}
In Fig.~\ref{fig:curve}, the image insets illustrate the intermediate completion results of the [0, 20, 40, 60, 80, 100]th step for one example depth image from the cabinet class. In addition,  $\mathcal{L}_{gt}(z)=\|GT-G({z};X)\|$ is averaged over all cabinet objects, where $GT$ is ground truth. For clarity, the curve is offset vertically by 0.2. $\Delta \mathcal{L}_{gt}(z) = \|GT-G(\mathring{z};X)\| - \|GT-G({z};X)\|$. We see the completed results are closer to ground truth than MVCN, though there is no ground truth supervision in inference. 

Fig.~\ref{fig:curve} illustrates empirically that, under the defined loss function, our optimization can find a good solution within 100 steps. The figure shows the average loss vs gradient descent steps on all the 150 cabinet test objects. 
We reach the maximum of $\mathcal{L}_{gen}(z)$
within $s$ steps, then the distance to $G(\mathring{z};X)$ decreases in the following $100-s$ steps. For 98\% of all the 1200 test objects, the maximum is reached within 10 steps ($s<10$), and within 20 steps for almost all. After 100 steps, the optimization has largely converged.

\subsection{Completion results}



\noindent \textbf{Improvements over Existing Works.} Here we compare our method with the state-of-the-art shape completion methods, including {3D-EPN} \cite{epn3d}, 	{FC} \cite{fc_Achlioptas2018LearningRA}, {Folding} \cite{folding}, three variants of  PCN \cite{ref_pcn}: PN2, PCN-CD, PCN-EMD, and MVCN \cite{mvcn}. TopNet \cite{topnet} is a recent point-based method, but their generated point clouds are sparse.


Table~\ref{tab:cmp_other} shows the quantitative results, where the completion results of the other methods are
from \cite{ref_pcn,mvcn} and `Direct-Opt' is the direct optimization method introduced above. With multi-view consistency optimization, both direct optimization and our method can improve MVCN on most categories of the test datasets, and our method achieves better results. The optimization methods fail on the Lamp dataset. As mentioned in \cite{mvcn}, the reason is that the completion of MVCN is bad on several lamp objects, which makes the optimization less meaningful.

Fig.~\ref{fig:cmp_mvcn} shows the qualitative improvements over the currently best view-based method, MVCN, where the normals of point clouds are color-coded. With the conditional generative net $G$ and multi-view consistency loss $C$, our method produces completed point clouds with smoother surfaces and fewer outliers, and can also fill holes of shapes on multiple categories.


\noindent \textbf{Completions results given different inputs}. Fig.~\ref{fig:diff_inputs} (a, b, c) show completed airplanes and cars under 3 different inputs of the same objects. Since the car input in (a) leaves a lot of ambiguity, the completed cars vary. The airplanes results are more similar because the inputs contain most of the structure.

\noindent \textbf{Multiple views of completed shapes.} Fig~\ref{fig:diff_inputs} (c) shows a completed airplane and car from 3 views. We see the completed shapes are consistent among  different views. 

\begin{figure}[h]
	\centering			
	\includegraphics[width=\linewidth]{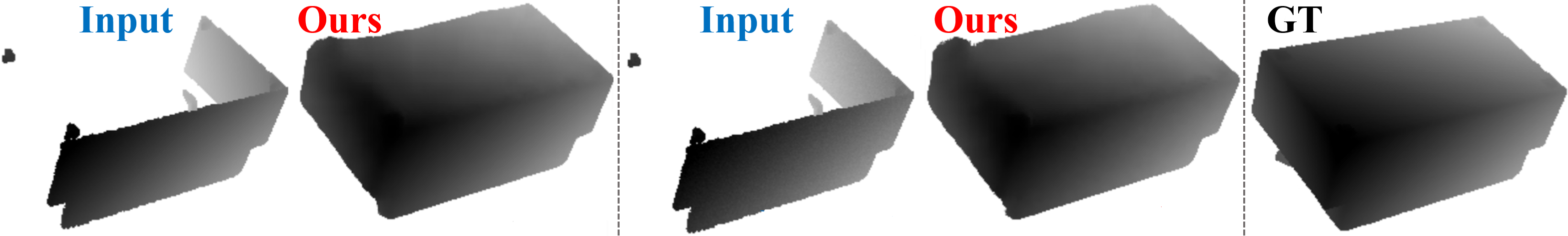}
	\captionof{figure}{Completions on noisy inputs. GT is ground truth.}
	\label{fig:noisy_input}
\end{figure}
\noindent \textbf{Completions on noisy inputs.}  In Fig.~\ref{fig:noisy_input}, we perturb the input depth map with Gaussian noise whose standard deviation is 0.01 times the scale of the depth measurements. Our completion is  robust to the noisy input.


\section{Conclusion}
\label{conclusion}
We proposed multi-view consistent inference to enforce geometric consistency in view-based 3D shape completion. We defined a novel multi-view consistency loss suitable for optimization in inference, which can be achieved without the supervision of ground truth. The experimental results demonstrate that our method can complete 3D shapes more accurately than existing methods. 


\begin{table*}[t]
	\begin{minipage}{\linewidth}
		\caption{Mean Chamfer Distance over multiple categories in ShapeNet. CD is scaled by $100$.}
		\vspace{-0.05in}
		\resizebox{\textwidth}{!}{
			\begin{tabular}{c|c|c|c|c|c|c|c|c|c}
				\hline
				Method & Avg & Airplane & Cabinet  & Car      & Chair    & Lamp                                                  & Sofa     & Table    & Vessel   \\
				\hline
				3D-EPN                                         & 2.0147                     & 1.3161                     & 2.1803                     & 2.0306                     & 1.8813                     & 2.5746                     & 2.1089                     & 2.1716                     & 1.8543                     \\
				FC                                         & 0.9799                     & 0.5698                     & 1.1023                     & 0.8775                     & 1.0969                     & 1.1131                     & 1.1756                     & 0.9320                     & 0.9720                     \\
				Folding                                    & 1.0074                     & 0.5965                     & 1.0831                     & 0.9272                     & 1.1245                     & 1.2172                     & 1.1630                     & 0.9453                     & 1.0027                     \\
				PN2                                        & 1.3999                     & 1.0300                     & 1.4735                     & 1.2187                     & 1.5775                     & 1.7615                     & 1.6183                     & 1.1676                     & 1.3521                     \\
				PCN-CD                                     & 0.9636                     & 0.5502                     & 1.0625                     & 0.8696                     & 1.0998                     & 1.1339                     & 1.1676                     & \textbf{0.8590 }                    & 0.9665                     \\
				PCN-EMD                                    & 1.0021                     & 0.5849                     & 1.0685                     & 0.9080                     & 1.1580                     & 1.1961                     & 1.2206                     & 0.9014                     & 0.9789                     \\
				MVCN                                       &           {0.8298}                   &{ 0.5273  }                   &{ 0.7154  }                   & {0.6322}                     & {1.0077 }                    & \textbf{1.0576  }                  & {0.9174 }                   & 0.9020         &{0.8790 } \\
				Direct-Opt                                     &           {0.8195}                   &{ 0.5182  }                   &{ 0.7001  }                   & {0.6156}                     & {0.9820 }                    & {1.1032}                  & {0.8885 }                   & 0.8854         &\textbf{0.8619} \\
				Ours                                       &           \textbf{0.8052}                   &\textbf{ 0.5175  }                   &\textbf{ 0.6722  }                   & \textbf{0.5817}                     & \textbf{0.9547 }                    & {1.1334  }                  & \textbf{0.8394 
				}                   & 0.8754                     &{0.8669}
		\end{tabular}}	
		\label{tab:cmp_other}
	\end{minipage}
	\begin{minipage}{\linewidth}
		\centering
		\vspace{0.06in}
		\includegraphics[width=\linewidth]{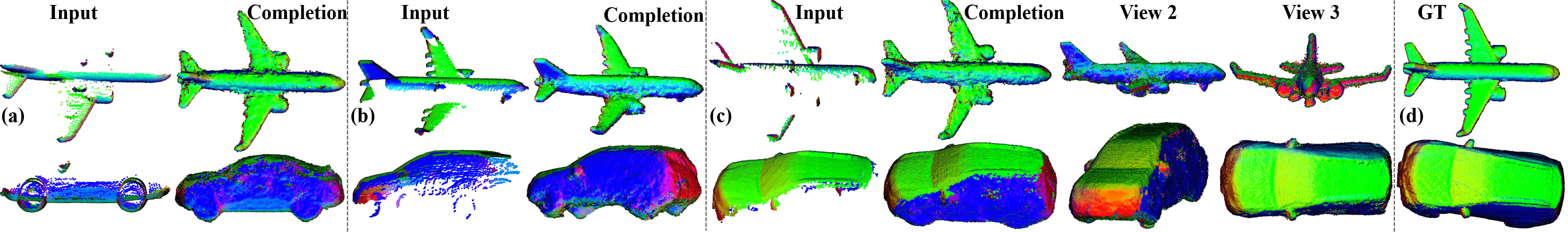}
		\captionof{figure}{Completion results given three different inputs (a, b, c). 3 different views (c). GT indicates ground truth (d).}
		\label{fig:diff_inputs}
	\end{minipage}
	\centerline{\begin{minipage}{0.78\linewidth}
			\vspace{0.06in}
			\begin{center}
				\includegraphics[width=\linewidth]{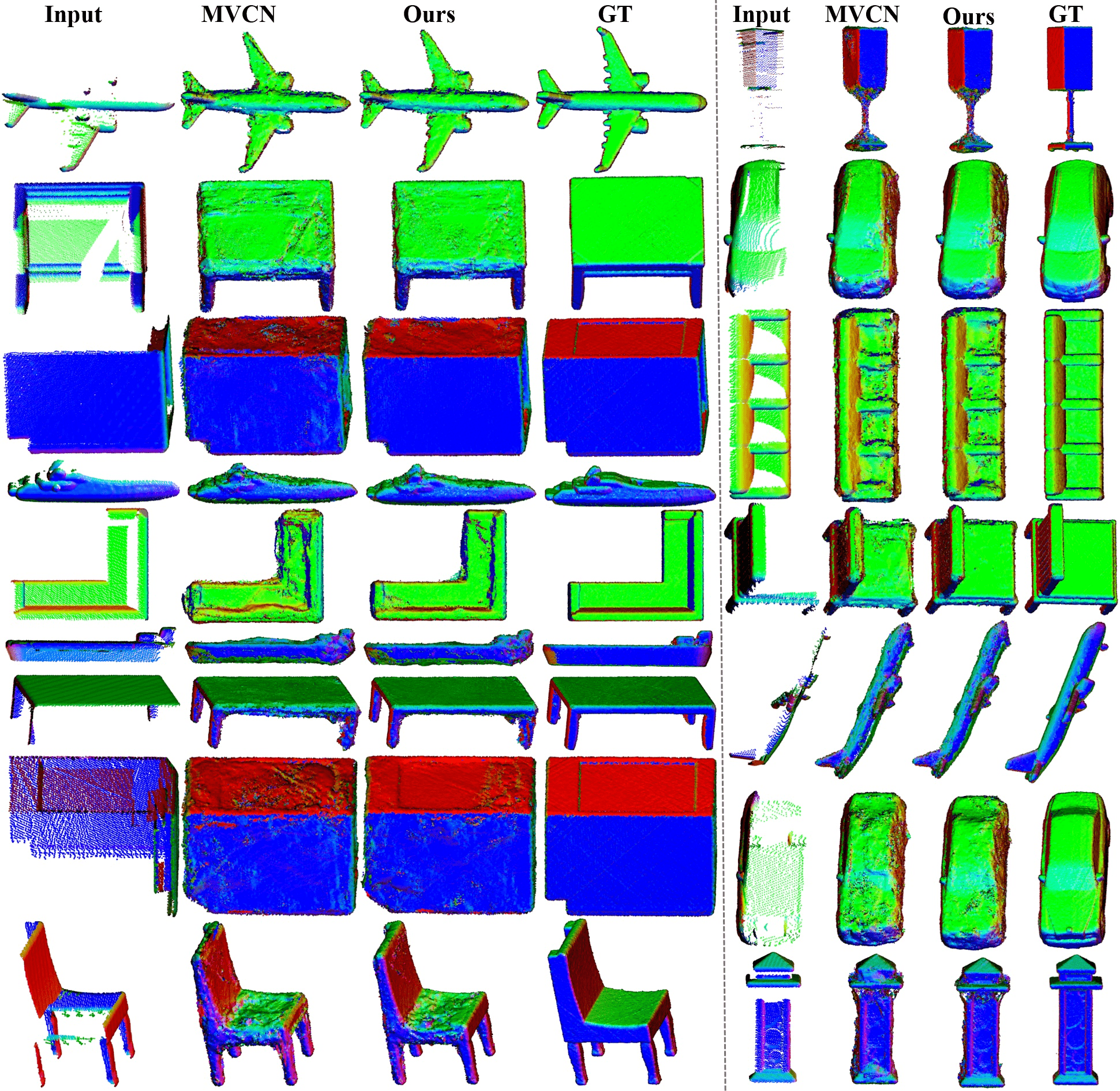}
				\captionof{figure}{Improvements over MVCN on multiple categories in ShapeNet. \textbf{GT} is ground truth.}
				\label{fig:cmp_mvcn}
			\end{center}
	\end{minipage}}
\end{table*}


\clearpage
{\small
	\bibliographystyle{aaai}
	\bibliography{egbib}
}





%
%

\end{document}